# Open-Vocabulary High-Resolution 3D (OVHR3D) Data Segmentation and Annotation Framework

**Jiuyi Xu, Meida Chen, Andrew Feng**
**USC Institute for Creative Technologies**
**Los Angeles, California**
{jiuxu, mechen, feng}@ict.usc.edu

**Zifan Yu**
**Arizona State University**
**Tempe, Arizona**
zifanyu@asu.edu

**Yangming Shi**
**Colorado School of Mines**
**Golden, Colorado**
yangming.shi@mines.edu

## ABSTRACT

In the domain of the U.S. Army modeling and simulation (M&S), the availability of high-quality annotated 3D data is pivotal to create virtual environments for training and simulations. Traditional methodologies for 3D semantic/instance segmentation, such as KpConv, RandLA, Mask3D, etc., are designed to train on extensive labeled datasets to obtain satisfactory performance in practical tasks. This requirement presents a significant challenge, given the inherent scarcity of manually annotated 3D datasets, particularly for the military use cases. Recognizing this gap, our previous research leverages the One World Terrain (OWT) data repository's manually annotated databases, as showcased at I/ITSEC 2019 and 2021, to enrich the training dataset for deep learning (DL) models. However, collecting and annotating large scale 3D data for specific tasks remains costly and inefficient.

To this end, the objective of this research is to design and develop a comprehensive and efficient framework for 3D segmentation tasks to assist in 3D data annotation. This framework integrates Grounding DINO (GDINO) and Segment-anything Model (SAM), augmented by an enhancement in 2D image rendering via 3D mesh. Furthermore, the authors have also developed a user-friendly interface (UI) that facilitates the 3D annotation process, offering intuitive visualization of rendered images and the 3D point cloud. To evaluate the proposed annotation framework, outdoor scenes from collected by using unmanned aerial vehicles (UAVs) and indoor scenes collected by using NavVis VLX and RGB-D camera in USC-ICT office building were used to conduct comparative experiments between manual methods and the proposed framework, focusing on 3D segmentation efficiency and accuracy. The results demonstrate that our proposed framework surpasses manual methods in efficiency, enabling faster 3D annotation without compromising on accuracy. This indicates that the potential of the framework to streamline the annotation process, thereby facilitating the training of more advanced models capable of understanding complex 3D environments with satisfactory precision.

## ABOUT THE AUTHORS

**Jiuyi Xu** is currently a master student in Computer Science from the University of Southern California. He is also a student research intern at USC-Institute for Creative Technologies (USC-ICT) supervised by Dr. Meida Chen. His research interests lie in 3D Object Detection and Segmentation. Email: jiuxu@ict.usc.edu

**Meida Chen** is currently a research scientist at USC-ICT. He received his Ph.D. degree from the University of Southern California. His research focuses on the applications of 3D computer vision for the creation of virtual environments training and simulation. He received the best paper award from the 2019 IITSEC Simulation Subcommittee and published numerous papers in top conferences/journals in both the computer vision and civil engineering fields. Email: mechen@ict.usc.edu

**Andrew Feng** is currently the Associate Director under the Geospatial Terrain Research group at USC-ICT. Dr. Feng holds a PhD and MS in Computer Science from the University of Illinois at Urbana-Champaign. He joined ICT in 2011 as a Research Programmer specializing in computer graphics, then as a Research Associate focusing on 3D avatar generation and motion synthesis, before becoming a Research Computer Scientist within the Geospatial Terrain Research group. Email: feng@ict.usc.edu

**Yangming Shi** is currently an assistant professor in Department of Civil and Environmental Engineering and a faculty member in the robotics program at the Colorado School of Mines. He received his Ph.D. from the University of Florida. His research focuses on Human-computer Interaction, Human-robot collaboration and NeuroErgonomics. Email: yangming.shi@mines.edu

**Zifan Yu** is currently a Ph.D. student in Computer Science from the Arizona State University supervised by Dr. Fengbo Ren. His research interests lie in computer vision, specifically 3D scene understanding such as photogrammetry and LiDAR point cloud semantic/instance segmentation. Email: zifanyu@asu.edu

# Open-Vocabulary High-Resolution 3D (OVHR3D) Data Segmentation and Annotation Framework

**Jiuyi Xu, Meida Chen, Andrew Feng**
**USC Institute for Creative Technologies**
**Los Angeles, California**
**{jiuxu, mechen, feng}@ict.usc.edu**

**Yangming Shi**
**Colorado School of Mines**
**Golden, Colorado**
**yangming.shi@mines.edu**

**Zifan Yu**
**Arizona State University**
**Tempe, Arizona**
**zifanyu@asu.edu**

## INTRODUCTION

The U.S Army and Navy have exhibited a strong interest in the field of rapid 3D reconstruction of battlefields, alongside virtual training and simulation tasks (Chen et al., 2020a). 3D data, particularly when annotated, is essential for these applications (Chen et al., 2020b,c,d). The use of unmanned aerial vehicles (UAVs) for the collection of aerial images and the application of photogrammetry techniques enables the swift collection and reconstruction of high-fidelity and geo-specific 3D terrain data (Chen et al., 2021, Hou et al., 2021). However, traditional manual annotation methods are considered both costly and time-intensive and the annotation of 3D data consistently presents significant challenges.

To this end, the authors developed an open-vocabulary high-resolution 3D (OVHR3D) data segmentation and annotation framework, leveraging the capabilities of Grounding DINO (GDINO) and Segment-Anything Model (SAM). Compared to traditional deep learning (DL) approaches, the proposed framework does not require the collection of 3D data for extensive training. It only necessitates the input of desired prompts and the selection of appropriate angles and altitudes which is required to render 2D images from the 3D mesh to be segmented. The framework automatically segments the rendered 2D images and back-projects the final labels to their corresponding positions in 3D point cloud. By rendering and segmenting from multiple different perspectives, the framework can facilitate the segmentation tasks for 3D point clouds and meshes. Additionally, to provide a more intuitive and user-friendly experience, a user interface (UI) was designed, grounded in the workflow of the proposed framework. The effectiveness of the framework was evaluated through a comparative study that measured the annotation performance using UI assistance against traditional manual annotation. This evaluation focused on the annotation accuracy, mIoU and consumed time, using ground truth labels which are obtained by annotation experts, across various indoor and outdoor environments – particularly indoor scenes from USC-ICT building and outdoor scenes collected with small UAVs. Results indicated a significant reduction in the annotation time with the new framework, while maintaining satisfactory accuracy and mIoU, highlighting its potential as a transformative tool for streamlining the process of 3D data annotation.

## BACKGROUND

In recent years, many or numerous DL models for 3D segmentation have been developed. These models use DL frameworks to automatically extract features within 3D data such as point clouds, meshes, or volumetric representations, thereby enabling accurate and efficient 3D segmentation tasks. Among these models, PointNet (Qi et al., 2017) is notable for its ability to classify and segment 3D point clouds into distinct groups, albeit with shortcomings in capturing local context. To overcome this, PointNet++ (Qi et al., 2017) improves upon the original by hierarchically organizing points and iteratively applying PointNet to these clusters, thereby enhancing the capture of local structures in point clouds. KpConv, or Kernel Point Convolution (Thomas et al., 2019), marks a further advancement by introducing a specialized convolution operation that processes point clouds directly without the need for a grid structure, greatly enhancing the extraction of local geometric features. Nonetheless, the computational intensity and recursive nature of these models impede their efficiency in practical applications. RandLA (Hu et al., 2019) presents a new method by utilizing random point cloud sampling to substantially decrease computational demands without sacrificing accuracy. Alongside point-based innovations, voxel-based models such as VoxelNet (Zhou & Tuzel, 2018) have emerged, converting point clouds into structured 3D voxel grids and combining the strengths of point and voxel processing to effectively handle sparse 3D data. SparseConv (Graham et al., 2018) further refines this approach by executing sparse convolutional operations that concentrate only on non-empty voxels, thus drastically reducing computational requirements and facilitating the processing of large-scale 3D data without compromising resolution or detail. With the success of the Transformer (Vaswani et al., 2017) architecture, Mask3D (Schult et al., 2022), which harnesses the Transformer structure, utilizes large-scale RGB-D datasets for self-supervised pre-training, embedding 3D structural knowledge into 2D feature maps without requiring complex 3D reconstructions or alignments. While these DL models demonstrate significant effectiveness in various 3D segmentation tasks, their dependency on extensive annotated 3D datasets for training continues to be a time-consuming and labor-intensive process.

The rapid evolution of DL models has also significantly advanced vision-language models (VLMs), which present innovative solutions to challenges of model adaptability and zero-shot prediction. VLMs, designed to interpret and produce content that merges visual and textual elements, introduce open-vocabulary methods. These methods enable models to detect and classify categories beyond predefined labels, substantially enhancing DL models' ability to process and understand unknown or rare words. This capability is crucial for adapting to open and dynamic environments, potentially addressing the challenge of limited annotated 3D data by circumventing the traditional model training process while maintaining the models' effectiveness. One key application of VLMs is open-vocabulary object detection (OVOD), firstly introduced in OVR-CNN (Zareian et al., 2021). By pre-training on extensive image-caption datasets, the vision encoder of OVR-CNN is able to acquire a more comprehensive visual-semantic correlation space, consequently enhancing its zero-shot detection ability. Recognizing the classification limitation of OVOD models in detection abilities compared to traditional DL object detection models, Detic (Zhou et al., 2022) proposes a novel strategy of jointly training the detection model using the ImageNet21K dataset (Ridnik et al., 2021) along with other object detection datasets to improve detection performance and expand category recognition. GLIP (Li et al., 2022) integrates OVOD with visual grounding, improving both OVOD and zero-shot object detection performance without relying on large-scale pre-training at the region-word level. The development of transformer-based detectors further advances this field, with DINO (Caron et al., 2021) establishing a foundation for powerful open-set detectors that incorporate multi-level textual data through grounded pre-training. The state-of-the-art (SOTA) model, Grounding DINO (GDINO) (Liu et al., 2023), combines DINO and GLIP to enable robust and adaptable detection of novel objects, bypassing extensive data acquisition and model training processes.

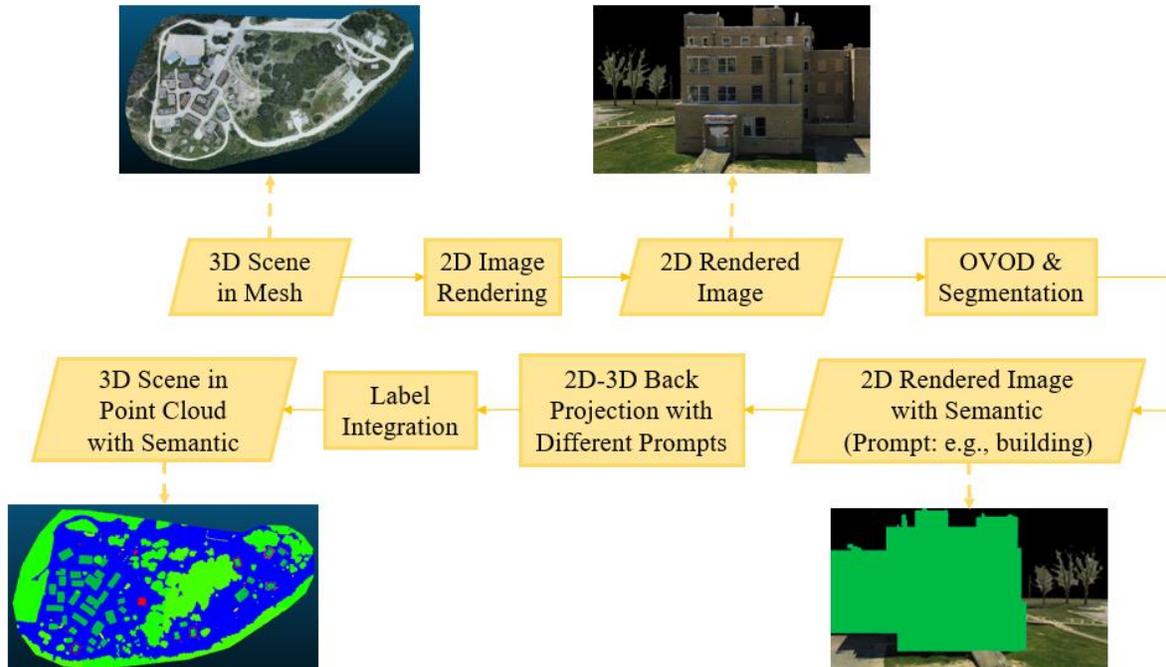

**Figure 1. Open-vocabulary High Resolution 3D Semantic Segmentation Framework**

In parallel, large language models (LLMs) pre-trained on extensive web text data are revolutionizing Natural Language Processing (NLP), demonstrating remarkable capabilities in zero-shot and few-shot generalization. These foundational models serve as a basis of a wide range of downstream NLP tasks. However, the exploration and development of foundation models in computer vision (CV) have been comparatively limited. Models like CLIP (Radford et al., 2021) and ALIGN (Jia et al., 2021) bridge the gap between vision and language, leveraging large-scale pre-training to achieve impressive generalization across various visual tasks. CLIP uses a contrastive learning framework to associate images and text, enabling tasks such as image classification and retrieval without task-specific training on labeled datasets. ALIGN aligns image and text representations through cross-modal learning, understanding semantic relationships between visual and textual inputs. Recently, the Segment-Anything-Model (SAM) (Kirillov et al., 2023) has marked a significant advancement in 2D segmentation, distinguished by its exceptional zero-shot inference capability. SAM emerges as an innovative foundation model for diverse CV applications, designed to process not only 2D images but also bounding boxes from object detection processes, thus integrating seamlessly with advancements in OVOD.

## THE FRAMEWORK FOR OVHR3D DATA SEGMENTATION AND ANNOTATION

The designed 3D data segmentation framework is illustrated in Figure 1. The framework consists of four major steps: (1) 2D image rendering, (2) 2D object bounding box detection, (3) fine object mask generation, and (4) 2D-3D back-projection and post-processing. In our framework, we initially render synthetic 2D images from 3D meshes using predetermined angles and camera parameters. Following that, we integrate GDINO and SAM to generate precise segmentation masks. Specifically, OVOD methods like GDINO facilitate straightforward identification of 2D object bounding boxes, while SAM enables fine-grained detection of object masks within these bounding boxes. Both semantic and instance annotations are derived using OVOD and segmentation foundation models within our designed pipeline. Following the segmentation of the rendered images, we perform 2D-3D back-projection along with subsequent post-processing steps, such as non-maximum suppression (NMS) (Hosang et al., 2017) and density-based spatial clustering of applications with noise (DBSCAN) (Ester et al., 1996), to project the final labels back onto the 3D point clouds.

### Rendering 2D Images From 3D Scenes

Due to the limited availability of aerial images captured by UAVs in the datasets used for the pre-trained models (e.g., GDINO and SAM), within our framework, direct detection and segmentation of aerial images by these models may not yield satisfactory results. In addition, the aerial images corresponding to 3D datasets often lack direct applicability for 2D object detection and segmentation because there are always critical issues such as data occlusion and object deformation with these images. To address these issues, leveraging 3D meshes becomes crucial since 3D meshes allow detailed rendering from different views with different camera poses. By rendering images slightly above ground elevation from reconstructed 3D meshes, we can generate images that closely resemble those in the training sets used for GDINO and SAM. This approach mitigates the difficulty of detection and segmentation tasks. Initially, we generate a corresponding 3D virtual scene from given 3D meshes. Subsequently, we adjust the intrinsic and extrinsic parameters of the virtual camera to render the desired 2D images effectively.

### Open-vocabulary Object Detection

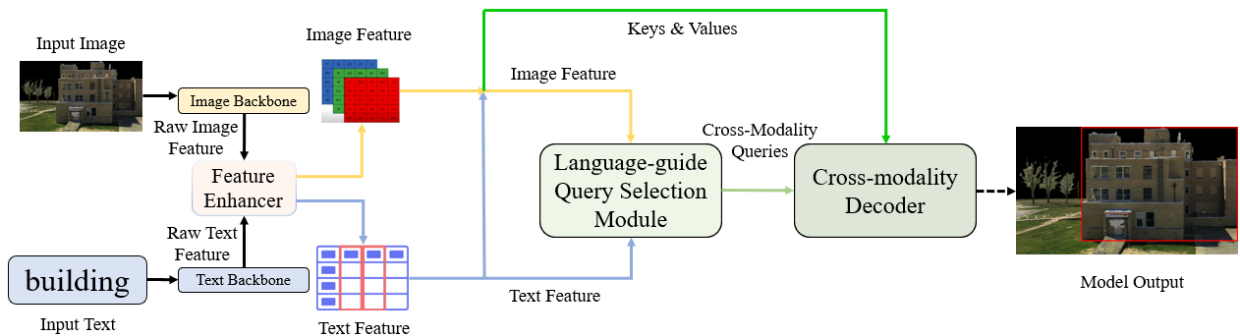

**Figure 2. Architecture of Grounding DINO**

After obtaining the rendered 2D images, we applied GDINO, an open-vocabulary detector, to generate bounding boxes for target objects. GDINO is a cutting-edge zero-shot object detection model that marries the powerful DINO architecture well-known for its success in image classification and object detection tasks with grounded pre-training.

Developed by IDEA-Research, GDINO can support natural language prompts as model inputs and can detect arbitrary objects based on human inputs, such as category names or referring expressions. In addition, it achieves the SOTA performance in open-set object detection tasks. A new feature in GDINO is the grounding module, which enhances the connection between language and visual content. During training, this module learns from a dataset containing images and corresponding text descriptions. By associating words in the text descriptions with specific regions within the images, GDINO is able to identify and detect objects in new images, even without prior knowledge of those objects. This capability significantly improves GDINO's performance in object detection tasks across diverse datasets and scenarios.

GDINO uniquely features a dual-encoder-single-decoder architecture, utilizing separate image and text encoders for feature extraction, a feature enhancement module for integrating these features, a language-guided query selection component for initializing queries, and a cross-modality decoder to refine detection boxes. Figure 2 illustrates GDINO's fundamental structure. For each image-text pair, GDINO first extracts fundamental image and text features using dedicated backbones. These foundational features are then fused in a feature enhancement module to create a unified cross-modality feature set. Employing a language-driven query selection mechanism, GDINO generates cross-modality queries based on the image features. Analogous to object queries in conventional Detection Transformer-based models, these queries undergo processing through a cross-modality decoder. This decoder interacts with both image and text features to iteratively refine and update the queries. Ultimately, the decoder's final layer produces queries used to identify object boxes and their corresponding textual descriptions.

**Segmentation Foundation Model**

SAM is another impressive model integrated into our framework to enhance the segmentation by generating precise masks based on customized prompts derived from detected target bounding boxes. SAM is a powerful foundation model that can perform well on new datasets for image segmentation tasks without large-scale labels and expensive training. SAM's advanced architecture enables it to adapt seamlessly to new image distributions and tasks without prior knowledge, a feature termed as zero-shot transfer. Trained on the extensive SA-1B dataset in a loop mode, which features over 1 billion masks across 11 million carefully selected images, SAM has demonstrated remarkable zero-shot performance, even surpassing previous benchmarks achieved through some fully supervised methods. During the training process, SAM not only learns from large-scale masks annotated by professionals but also continually refines its capabilities by generating and incorporating new masks autonomously.

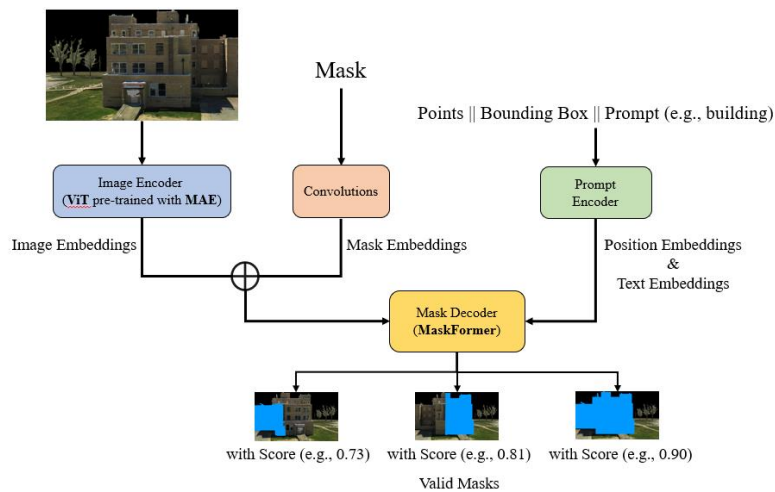

**Figure 3. Architecture of Segment-Anything Model**

SAM consists of three core components: an image encoder leveraging a pre-trained Vision Transformer (ViT) (Dosovitskiy et al., 2020) adapted for handling high-resolution inputs efficiently, a versatile prompt encoder for interpreting sparse (points, boxes, text) and dense (masks) prompts, and a mask decoder responsible for generating masks based on combined image and prompt embeddings, as shown in Figure 3. This unique architecture enables flexible prompting, real-time mask computation, and ambiguity awareness in segmentation tasks. The prompt encoder handles different prompt types through innovative positional encodings and convolutional embedding techniques. The mask decoder utilizes a modified Transformer decoder block to integrate image and prompt embeddings, facilitating bi-directional interaction. SAM predicts multiple masks for ambiguous prompts and employs a dynamic linear classifier to assign foreground probabilities at each pixel. Additionally, it ranks masks based on confidence scores (estimated IoU), ensuring optimal segmentation results.

**Back-projection And Post-processing Approaches**

Depth map and corresponding camera pose enable straightforward projection/back projection between RGB-D images and 3D data. This capability is crucial for tasks where accurate alignment and labeling between 2D images and 3D point clouds are required. In our framework, this process leverages depth maps generated during 2D image rendering from 3D mesh and the corresponding camera pose to accurately map points from the 2D image space to the 3D virtual scene. For each pixel in the depth map, we transform the 2D pixel coordinates to the 3D space using the camera's intrinsic and extrinsic parameters. Consequently, the 2D annotations can be back-projected to the 3D space.

Following the completion of the 2D-3D back-projection process in our framework, additional post-processing techniques are employed to refine the noise and inconsistency for the annotated 3D points caused by the nature of muti-view projections. These approaches, including NMS and connected component algorithms. NMS has been widely used in CV, addresses the issue of multiple overlapping entities in 2D instance segmentation task. It selectively retains the most prominent entity while suppressing redundant or overlapping annotations. This ensures that each object or entity in the 2D image is represented by a single, optimal bounding box, thereby improving the clarity and reliability of subsequent analysis and applications. In our framework, we implemented a 3D NMS algorithm to improve the performance of the proposed framework. This 3D NMS algorithm processes a set of 3D semantic objects to remove redundant or overlapping instances, keeping only the most significant ones based on confidence scores. It starts by sorting the objects by their confidence. Following that, it selects the object with the highest confidence and checks all other objects for significant overlap. If two objects overlap beyond a certain threshold, they are merged. This process repeats until all objects have been checked and only non-overlapping ones remain.

Moreover, during the 2D detection and segmentation tasks using GDINO and SAM, it is common to output inaccurate masks, such as over-segmentation of pixels along the edges of objects like houses. This can lead to mis-projections in the 3D point cloud, where these pixels may inaccurately project onto nearby object such as ground instead of aligning with the true object boundaries. To address this challenge, especially in scenarios involving multiple viewpoints of the same object, applying a connected component algorithm is particularly effective during the 2D-3D back-projection process since it can eliminate incorrect segmented points that are projected significantly distant from the actual house in 3D space. One such algorithm, DBSCAN, stands out for its ability to identify and merge 3D points of the same object across different perspectives into coherent clusters. This approach not only corrects mis-projections but also enhances the overall accuracy and consistency of annotations within our framework. Together, these post-processing techniques—NMS for handling redundant 3D semantic instances and DBSCAN for correcting mis-projected annotations across viewpoints—ensure that the annotated 3D data are refined and optimized for subsequent analysis, visualization, and modeling tasks within our proposed framework. These methods not only improve the accuracy of annotations but also contribute to the robustness and reliability of the entire 3D data processing pipeline.

**EXPERIMENTS AND RESULTS**

In this section, we present an evaluation of the proposed open-vocabulary framework for high-resolution 3D data segmentation and annotation. We conducted the experiments to evaluate the efficiency of the proposed framework in comparison to traditional manual segmentation. We also provide a detailed discussion of the experimental results to better demonstrate the advantages of our proposed framework.

**Datasets and Evaluation Metrics**

We evaluated the proposed framework in both indoor and outdoor environments. The indoor dataset we used are two USC-ICT office scenes (i.e., Event Room and Lab) obtained by using NavVis VLX. The outdoor dataset (i.e., Elijah dataset) was captured with Parrot Anafi. The 3D meshes were reconstructed using ContextCapture. The reconstructed 3D meshes are illustrated in Figure 4.

In the indoor environment, we evaluated the performance of the proposed framework on 12 objects, including painting (1), wall (2), door (3), ceiling (4), light (5), window (6), trash can (7), table (8), chair (9), floor (10), shelf (11), and monitor (12). In the outdoor environment, we evaluated the proposed framework on 8 objects, including ground, building, tree, clutter, vehicle, pole, fence, and special object. We used mean Intersection over Union (mIoU) and accuracy as our primary evaluation metrics. Additionally, to highlight the advantages of the proposed framework, we also included the time consumed in segmentation task as an auxiliary metric in our experiments.

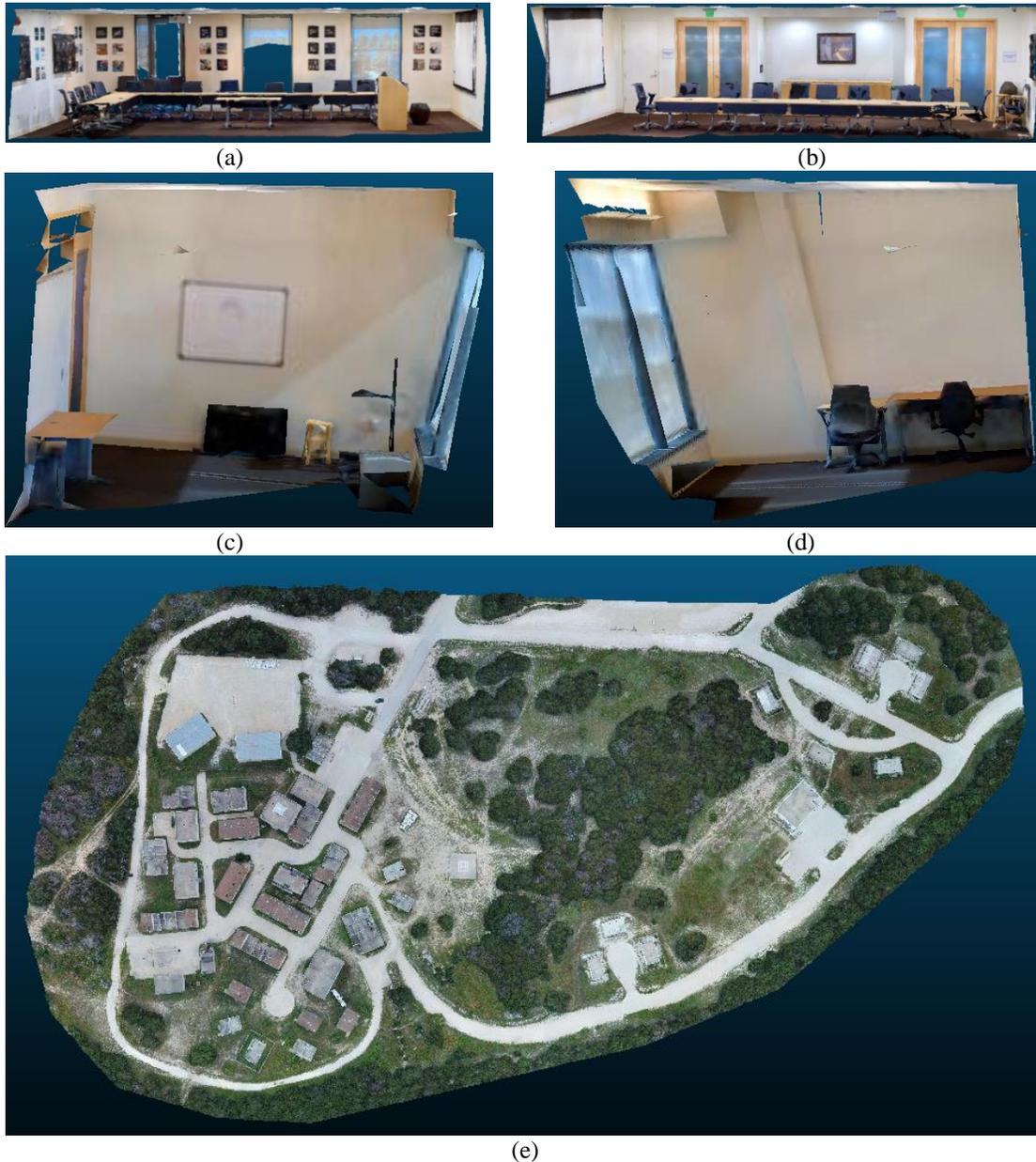

**Figure 4. Indoor and outdoor datasets: (a) and (b) the Event Room scene; (c) and (d) the Lab scene; and (e) the U.S. Army collected outdoor scene**

**Results and Analysis**

In the experiment, two 3D annotation experts initially annotated the selected indoor and outdoor datasets to establish the ground truth. Following that, the experts segmented the same datasets using our designed OVHR3D (the proposed framework) UI tool. We then conducted a detailed comparison between the UI-assisted annotation results and the fully manual annotation results. The result of the indoor setting, presented in terms of mIoU and accuracy, is shown in Table 1. The result of the outdoor setting, also presented in terms of mIoU and accuracy is shown in Table 2. As the supplementary experimental result, the time consumption for the experiments is summarized in Table 3.

**Table 1. Segmentation Results of the Indoor Setting**

| Dataset | Accuracy (%) | mIoU (%) | Per Class mIoU (%) | | | | | | | | | | | |
|---|---|---|---|---|---|---|---|---|---|---|---|---|---|---|
| | | | 1 | 2 | 3 | 4 | 5 | 6 | 7 | 8 | 9 | 10 | 11 | 12 |
| | 95.7 | 93.6 | 88.5 | 88.7 | 87.5 | 95.7 | 91.9 | 95.4 | 88.3 | 90.5 | 82.2 | 72.2 | 86.4 | 87.8 | 95.3 |
| | 94.8 | 94.9 | 89.2 | 92.8 | 88.8 | 94.8 | 90.1 | 100 | 90.5 | 91.7 | 77.4 | 74.6 | 91.2 | 84.8 | 93.3 |

**Table 2. Segmentation Results of the Outdoor Setting**

| Dataset | Accuracy (%) | mIoU (%) | Per Class mIoU (%) | | | | | | | |
|---|---|---|---|---|---|---|---|---|---|---|
| | | | ground | building | tree | clutter | vehicle | pole | fence | special object |
| Tile1 | 94.70 | 89.26 | 88.04 | 97.23 | 90.52 | 83.60 | - | - | 79.91 | 96.25 |
| Tile2 | 94.98 | 83.99 | 89.98 | - | 90.88 | 85.48 | - | - | 73.72 | 79.91 |
| Tile3 | 96.43 | 90.77 | 95.28 | 98.05 | 86.67 | 85.13 | - | - | 88.72 | - |
| Tile4 | 94.26 | 90.70 | 92.83 | 97.00 | 88.43 | 75.34 | - | 97.89 | 91.98 | 91.42 |
| Tile5 | 98.04 | 90.00 | 96.73 | 97.72 | 87.57 | 89.84 | 90.20 | 94.84 | 75.38 | 87.75 |
| Tile6 | 94.15 | 87.86 | 87.51 | - | 90.09 | 82.83 | - | - | - | 91.02 |
| Tile7 | 96.36 | 92.56 | 84.69 | 98.40 | 95.38 | - | - | 96.31 | 88.03 | - |
| Tile8 | 96.48 | 88.62 | 94.67 | 97.66 | 88.64 | 90.36 | - | 95.65 | 75.52 | 77.78 |
| Tile9 | 96.42 | 92.52 | 90.05 | 96.51 | 86.60 | - | - | - | - | 96.71 |
| Tile10 | 95.59 | 91.67 | 93.42 | 95.43 | 83.39 | - | - | - | - | 94.45 |
| Tile11 | 97.40 | 95.68 | 94.26 | 94.74 | 95.33 | - | - | - | - | 98.40 |
| Tile12 | 98.82 | 95.83 | 91.12 | 96.57 | 91.98 | 96.64 | 98.88 | 98.96 | 96.65 | - |
| Tile13 | 98.21 | 96.31 | 95.51 | - | 97.12 | - | 96.30 | - | - | - |
| Tile14 | 98.05 | 92.53 | 96.40 | 97.53 | 95.95 | - | 80.67 | - | 92.08 | - |
| Tile15 | 92.99 | 89.49 | 84.99 | 98.08 | 86.70 | - | 95.25 | - | - | 82.43 |
| Tile16 | 96.74 | 94.54 | 93.35 | 99.48 | 93.81 | - | - | 95.26 | 90.74 | - |

In Table 1, the accuracy and mIoU metrics are satisfactory for both indoor scenes, indicating that the proposed framework performs well in these environments. The slight variation between the Event Room and Lab suggests that the framework is consistent across different indoor settings. The per-class mIoU values vary, with some classes like "door" and "light" achieving above 95%, while others like "trash can" and "monitor" show lower values around the mid-70s to low-80s percentage. The consistent performance across most classes highlights the robustness of the framework in identifying various indoor objects. In Table 2, the outdoor dataset results show that the framework maintains a high level of accuracy and mIoU across a variety of scenes and objects, though certain classes show more variability in performance. The accuracy metric ranges from 92.99% (Tile15) to 98.82% (Tile13), demonstrating high accuracy across different outdoor scenes. Meanwhile, the mIoU metric ranges from 83.99% (Tile2) to 92.77% (Tile10), indicating that the framework can effectively segment outdoor scenes as well. The "building" and "special object" classes consistently achieve high mIoU values (above 95% in several tiles). Some classes like "clutter" and "fence" show more variability, with values ranging from the mid-70s to mid-90s. In Table 3, the results clearly show that the proposed UI-assisted framework significantly reduces the time required for annotation. This substantial reduction in annotation time demonstrates the efficiency and practicality of the proposed framework, making it a valuable tool for large-scale 3D data segmentation tasks. The experimental results suggest that the proposed open-vocabulary framework for high-resolution 3D data segmentation and annotation is highly effective in both indoor and outdoor environments. Figure 5 shows an example where our UI performs very well in annotating complex objects, such as "fence". It achieves high accuracy and mIoU values across different scenes and object classes, and significantly reduces the time required for manual annotation, thereby increasing efficiency and productivity in 3D data segmentation tasks.

Compared to traditional manual 3D data annotation methods, we no longer need to spend significant effort in finding object edges and carefully annotating them. Moreover, our open-vocabulary framework can complete the detection and segmentation of rendered images within a short period (within seconds), thus significantly reducing annotation time. By employing the assistive UI, the framework automatically segments the rendered images based on the input prompt. Given that the detection and segmentation capabilities of GDINO and SAM have been proved, the segmentation results are satisfactory. Considering that manual annotation is prone to errors, especially for more complex 3D data, the performance of our model should be better, as our ground truth is based on human-annotated data. Figure 6 shows an example that our UI help with annotation better than traditional manual methods. From the result, our UI demonstrates more logical object boundary annotations and is more accurate compared to manual annotations. Despite our framework's overall effectiveness, the experiment results also show the inconsistent performance in segmenting different objects in the same environment and in segmenting the same objects in different environments. We attribute this to the data quality difference caused by the data collection.

**Table 3. Summary of Consumed Time**

| Dataset | Fully Manual Exp (minutes) | UI-Assisted Exp (minutes) |
|---|---|---|
| Tile1 | 44 | 16 |
| Tile2 | 37 | 13 |
| Tile3 | 31 | 11 |
| Tile4 | 40 | 14 |
| Tile5 | 47 | 16 |
| Tile6 | 15 | 7 |
| Tile7 | 9 | 5 |
| Tile8 | 50 | 18 |
| Tile9 | 68 | 25 |
| Tile10 | 70 | 25 |
| Tile11 | 31 | 11 |
| Tile12 | 39 | 15 |
| Tile13 | 7 | 4 |
| Tile14 | 56 | 20 |
| Tile15 | 57 | 24 |
| Tile16 | 11 | 5 |
| Event Room | 221 | 86 |
| Lab | 88 | 37 |

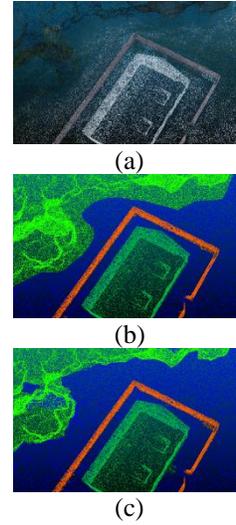

(a)

(b)

(c)

**Figure 5. Experiment Result Visualization: (a) RBG Image, (b) Ground Truth, (c) UI-Assisted Annotation**

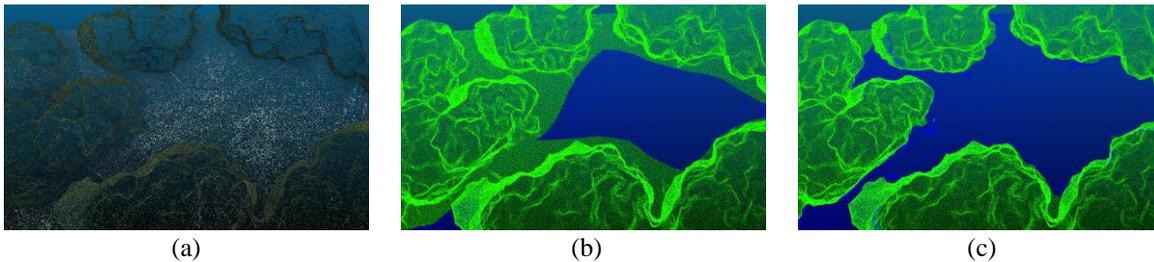

(a)                                    (b)                                    (c)

**Figure 6. Experiment Result Visualization: (a) RBG Image, (b) Ground Truth, (c) UI-Assisted Annotation**

## APPLICATIONS

This research has presented a CV-oriented open-vocabulary framework designed to annotate 3D data. This experiment results demonstrate the framework's efficiency and potential in creating high-quality 3D annotated datasets across various settings. The significance of this framework is particularly pronounced in the realm of U.S. Army Modeling and Simulation (M&S), where it plays a pivotal role in streamlining the annotation of complex 3D data. By optimizing this process, the framework minimizes reliance on manual labor, thereby boosting overall efficiency and productivity. This is achieved through precise segmentation of objects such as terrain features and indoor targets across a spectrum of real-world and simulated environments. In practical terms, the framework's ability to annotate 3D data with an open vocabulary not only enhances accuracy but also expands the scope of applications in military simulations. It facilitates robust target tracking and scene understanding, critical for operational planning and training scenarios. By automating and refining the annotation process, the framework empowers military personnel to focus on higher-level tasks, ultimately enhancing decision-making capabilities and operational readiness. Moreover, the framework's adaptability to various settings underscores its versatility and potential for broader applications beyond military simulations.

Industries ranging from architecture and urban planning to robotics and autonomous systems can benefit from its capabilities in creating comprehensive and reliable 3D datasets.

## LIMITATIONS AND FUTURE WORK

While our proposed framework has shown promising performance in experiments, several limitations exist. First, the efficacy of the framework hinges significantly on the quality of 3D reconstruction, which can be influenced by variables such as varying weather conditions and dynamic landscapes. Therefore, it is crucial to validate the framework across a broader range of diverse and complex scenarios to ensure its robustness and reliability for U.S. Army M&S applications. Furthermore, the accuracy of annotation results is closely related to the quality of the 2D rendered images, which is influenced by factors like the UI settings during image capture. Future research efforts could benefit from a more comprehensive exploration to determine the optimal angle and altitude for image capture, thereby enhancing the overall efficiency and effectiveness of the open-vocabulary framework. This optimization could involve refining parameters related to camera positioning and settings to maximize the detail captured in 2D images, thereby improving subsequent 3D reconstruction and annotation processes. Additionally, exploring the application of neural radiance rendering to enhance the realism of the rendered 2D images represents another avenue for future investigation. Integrating advanced rendering methods such as 3D Gaussian could potentially mitigate artifacts and improve the fidelity of rendered images, thereby contributing to more accurate and reliable annotations in complex environments.

## CONCLUSION

This research introduces a semi-automated annotation framework that leverages OVOD and advanced segmentation model to label 3D point clouds across diverse indoor and outdoor environments. The framework comprises a structured pipeline encompassing four main stages: (1) rendering 2D synthetic images from 3D meshes based on predetermined angles and camera parameters which establishes the foundational images required for subsequent annotation processes, (2) detecting 2D object bounding boxes utilizing GDINO which identifies and localizes objects within the rendered images, laying the groundwork for detailed segmentation, (3) generating fine object masks by employing SAM, and (4) performing 2D-3D back-projection and post-processing to refine the segmentation results. To evaluate the performance of this framework, comprehensive experiments were conducted across several distinct scenarios encompassing indoor and outdoor environments. Indoor scenarios utilized scanned data derived from NavVis VLX, while outdoor environments were based on data collected by Parrot Anafi. The results demonstrate the framework's effectiveness in annotating a wide range of objects. The capability of the proposed framework to annotate objects accurately in military-related scenes suggests its potential in automating tasks in the field of U.S. Army M&S. By reducing manual effort and enhancing annotation efficiency, the framework can significantly contribute to improving human operation and decision-making within military modeling and simulation applications.

## ACKNOWLEDGEMENTS

The authors would like to thank our primary sponsors of this research: Mr. Clayton Burford of the Battlespace Content Creation (BCC) team at Simulation and Training Technology Center (STTC). This work is supported by University Affiliated Research Center (UARC) award W911NF-14-D-0005. Statements and opinions expressed and content included do not necessarily reflect the position or the policy of the Government, and no official endorsement should be inferred.